\documentclass[letterpaper, 10 pt, conference]{ieeeconf}
\IEEEoverridecommandlockouts
\IEEEtriggeratref{12} 

\usepackage{xcolor}
\usepackage{amssymb} 
\usepackage{amsmath}
\usepackage{subcaption}
\usepackage{float}
\usepackage{booktabs}
\usepackage{graphicx}
\usepackage{tikz} 
\usetikzlibrary{arrows,decorations.markings}
\usepackage{cite}
\usepackage{authblk}

\title{\LARGE \bf
MoboTSP: Solving the Task Sequencing Problem for\\ Mobile Manipulators}

\author{Nicholas Adrian and Quang-Cuong Pham%
  \thanks{The authors are with the HP-NTU Digital Manufacturing
  Corporate Lab and School of Mechanical and Aerospace Engineering,
  Nanyang Technological University, Singapore. Email:
  nicholasadr@ntu.edu.sg}}

\begin{document}

\maketitle
\thispagestyle{empty}
\pagestyle{empty}

\begin{abstract}
  We introduce a new approach to tackle the mobile manipulator task
  sequencing problem. We leverage computational geometry, graph theory
  and combinatorial optimization to yield a principled method to
  segment the task-space targets into clusters, analytically
  determine reachable base pose for each cluster, and find task
  sequences that minimize the number of base movements and robot
  execution time. By clustering targets first and by doing so from
  first principles, our solution is more general and computationally
  efficient when compared to existing methods.
\end{abstract}

\section{Introduction}
\label{part:introduction}

A  mobile manipulator consists of a manipulator (e.g. a robotic
arm) mounted unto a mobile base. The combined system extends the
workspace of a fixed-based manipulator \cite{vafadar2018optimal}\cite{xu2020}\cite{shin2003motion}.

\begin{figure}[!t]
  \centering
  \includegraphics[width=\columnwidth]{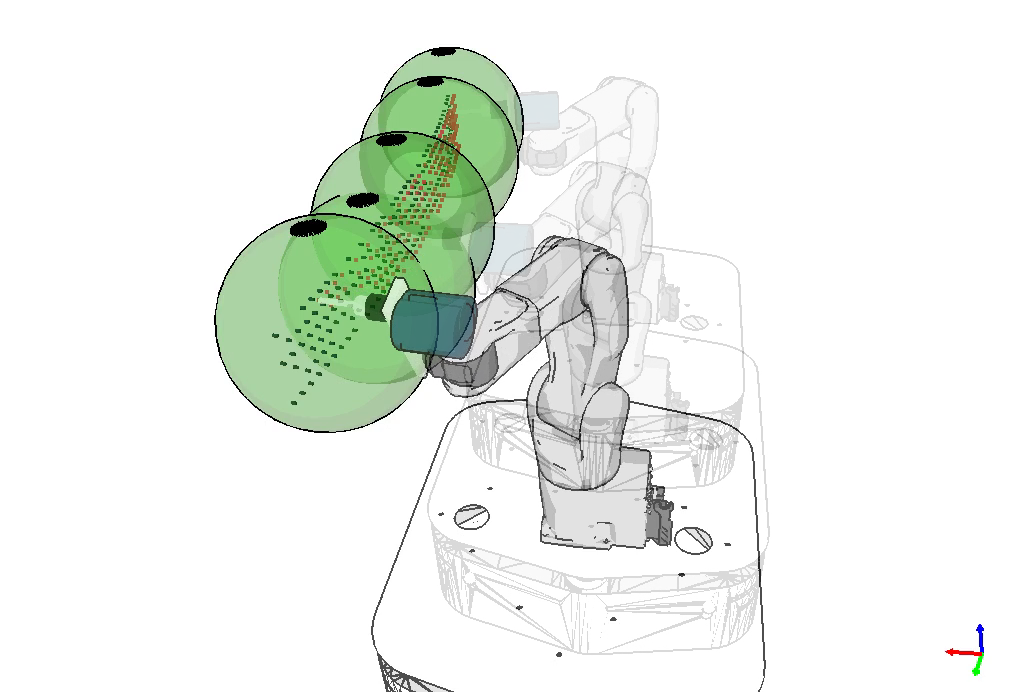}
  \caption{Example of a mobile manipulator consisting of a Denso VS60
    robot arm mounted on a a Clearpath Ridgeback mobile base. The task
    is to perform drilling on a set of targets on a curved
    surface. There is no \emph{a priori} constraint on the target
    cluster or sequence.}
  \label{fig:world}
\end{figure}

A frequent task for a mobile manipulator is to visit a set of ordered
or unordered targets scattered around the workspace: think for example
of drilling multiple holes on aircraft fuselages or on curved housing
walls (Fig.~\ref{fig:world}). Another example is motivated by mobile
3D-printing of large workpieces~\cite{efe2019}.

For fixed-based manipulators, finding the optimal sequence of targets
and the corresponding Inverse Kinematics (IK) solutions is known as
the Robotic Task Sequencing Problem (RTSP)
see~\cite{suarez2018robotsp} for a recent review. Compared to
fixed-base manipulators, solving RTSP for mobile manipulators comes
with extra complexities due to the distribution of task space targets
that spans beyond the reachable workspace of a fixed-base
manipulator.

One solution may consist of segmenting \emph{a priori} the targets
into what we call \textit{workspace clusters}. Each \textit{workspace
  cluster} would contain \textit{workspace targets} that can be
reached by a corresponding \textit{workspace base pose}. However, when there is no evident of \emph{a priori} spatial clusters, this
hierarchical approach will likely result in suboptimal solutions.

Here we introduce a new approach to tackle mobile manipulator task
sequencing problem. We leverage computational geometry, graph theory
and combinatorial optimization to yield a principled method to segment
the task point targets into clusters, analytically determine reachable
base pose for each cluster, and find task sequences that minimize the
number of base movements and robot execution time.

Note that the minimization of the number of base movements has a high priority, as such movements are the largest source of
localization errors (as compared to the kinematic errors of a
fixed-base manipulator).

Our contributions are as follows:
\begin{itemize}
\item A method to select a volume (represented as set of balls) in the manipulator's reachable workspace which will be sufficient to reach all the targets;
\item A method to cluster unordered targets into \emph{workspace clusters} consisting of \emph{workspace targets} and reachable \emph{workspace base pose} in each cluster;
\item A task sequencing algorithm for mobile manipulator, minimizing
  the number of base movements and the manipulator trajectory time
  within and between \textit{workspace clusters}.
\end{itemize}

The remainder of the paper is organized as follows. In Section
\ref{part:lit-review}, we review related works in mobile manipulator
task sequencing. In Section \ref{part:pipeline}, we introduce the
overall pipeline. In Section \ref{part:closed-balls}, we present the
detailed calculations of the set of balls. In Section
\ref{part:covering}, we introduce our method to cluster the targets
into separate workspaces, followed by task sequencing algorithm in
Section \ref{part:task-sequencing}. We present the simulation results
in Section \ref{part:experiment}. Finally, in Section
\ref{part:conclusion}, we conclude and sketch directions for future
work.

\section{Literature Review}
\label{part:lit-review}

To start with, we look at some works related to finding robot base placement. In \cite{zacharias2007capturing}, reachability map is introduced as a 3D workspace representation that represents the reachability probability of points in task space. The reachability map is used in \cite{zacharias2008positioning} to find robot placement for a constrained linear trajectory. The work is extended in \cite{zacharias2009using} to 3D trajectory. In \cite{vahrenkamp2012manipulability}, extended manipulability measure as quality index is used as precomputed workspace representation instead of reachability. Robot placement can also be found using Inverse Reachability Distribution \cite{vahrenkamp2013robot}.

Several works on RTSP have been attempted but they are mostly limited to fixed-base manipulator. In \cite{zacharia2005optimal}, Genetic Algorithm (GA) is employed on 3-DOF and 6-DOF robots with multiple IK solutions per target. They reported 1800 seconds CPU time for 50 task points. This work was extended in \cite{baizid2010genetic} to include base placement of the robot. No computational details were provided for varying number of task points or base permutations per task point. In \cite{saha2006planning}, the author considered 50 targets with five configurations each and the solution was found in 9621.64 seconds. The fastest solution for fixed-based manipulator RTSP, to our best understanding, is found in \cite{suarez2018robotsp}. A fast near-optimal solution was found using a three-step algorithm exploiting both task and configuration space. In their experiment, they visited 245 targets with an average of 28.5 configurations per target and the solution was found in 10 seconds with 60 seconds execution time. For the case of mobile manipulator, a search on 10-DOF configuration space is done as part of Genetic Algorithm-based optimization to minimize overall configurations displacement in \cite{vafadar2018optimal}.

In \cite{xu2020}, a relatively similar problem of part-supply pick and place is presented where the robot had to pick parts from multiple tray locations. Given \textit{m} number of trays containing \textit{n} number of objects with \textit{o} number of possible grasping poses each, $m\times n\times o$ IK queries would be required. Objects are also pre-clustered into trays which reduce the number of base regions to consider. However, such target clustering might not be available or easily defined in other cases such as when continuous target distribution is involved.

In \cite{shin2003motion}, the mobile manipulator's end-effector has to follow a continuous path trajectory. To deal with inaccurate base locomotion, the author used manipulability performance function to find minimum base stopping positions that will allow the robot to follow discretized points on the path trajectory. However this method requires the discretized target points to be pre-sequenced. 

Compared to other methods, our approach exploits the often overlooked strategy of clustering targets first. This allows us to have a fast and computationally efficient solution. In Section \ref{part:experiment}, we show that we segmented 183 targets into clusters, while also finding the respective base poses, in just 0.104 seconds. To our best understanding, there is no other systematic solution available to the mobile manipulator task sequencing problem which can deal with indeterminate target segmentation and has fast computation time.

\section{Pipeline}
\label{part:pipeline}

We assume that the targets' positions $\boldsymbol{X} = [\boldsymbol{x}_1, ..., \boldsymbol{x}_n] \in \mathbb{R}^{n\times3}$ and their respective drilling direction vectors $\boldsymbol{R} \in \mathbb{R}^{n\times3}$ are provided for all $n$ targets.

\begin{figure}[!ht]
  \centering
  \begin{subfigure}{0.5\textwidth}
    \includegraphics[width=1\linewidth]{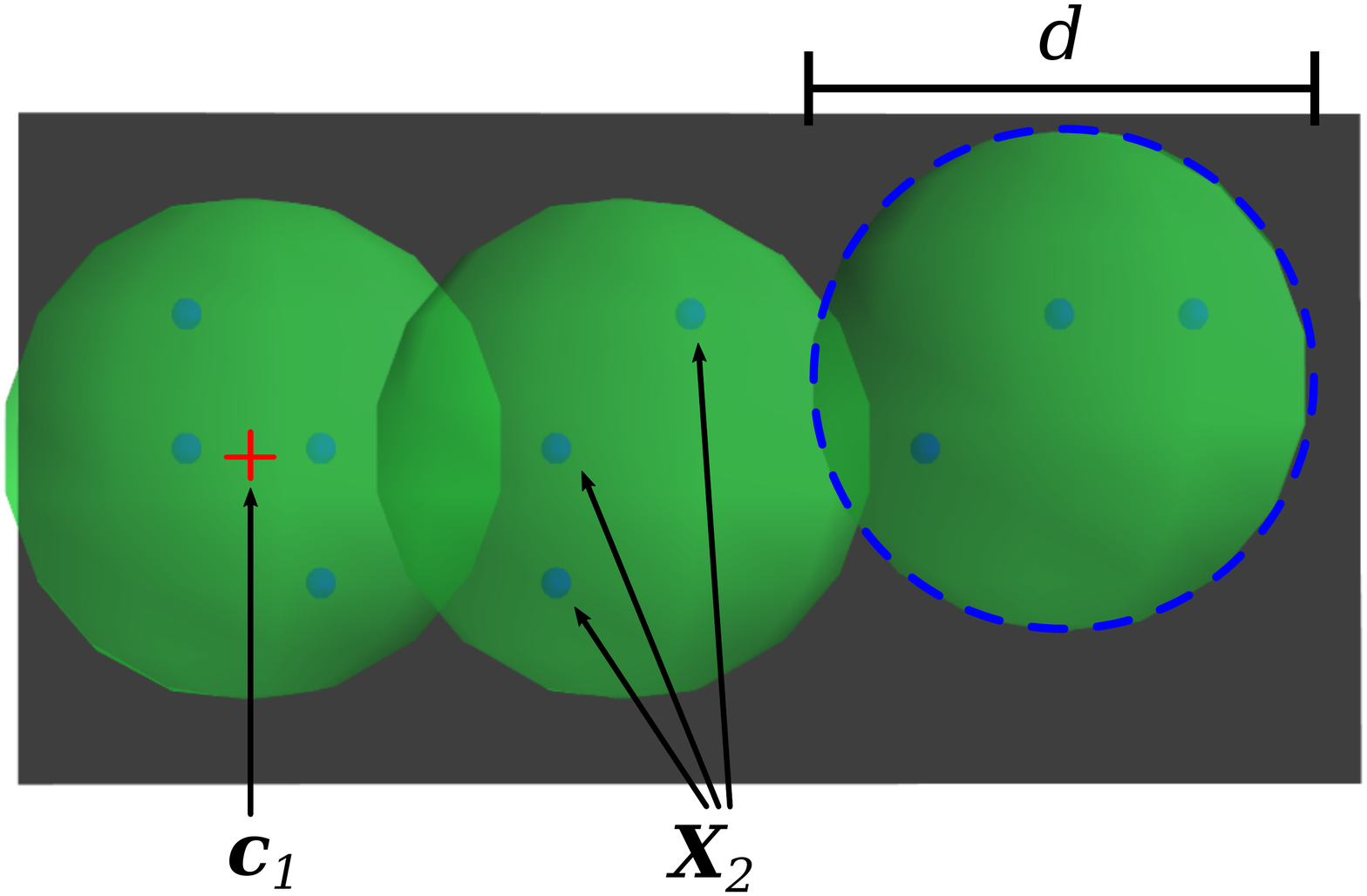}
    \caption{}
    \label{fig:covering}
  \end{subfigure}
  \begin{subfigure}{0.5\textwidth}
    \includegraphics[width=1\linewidth]{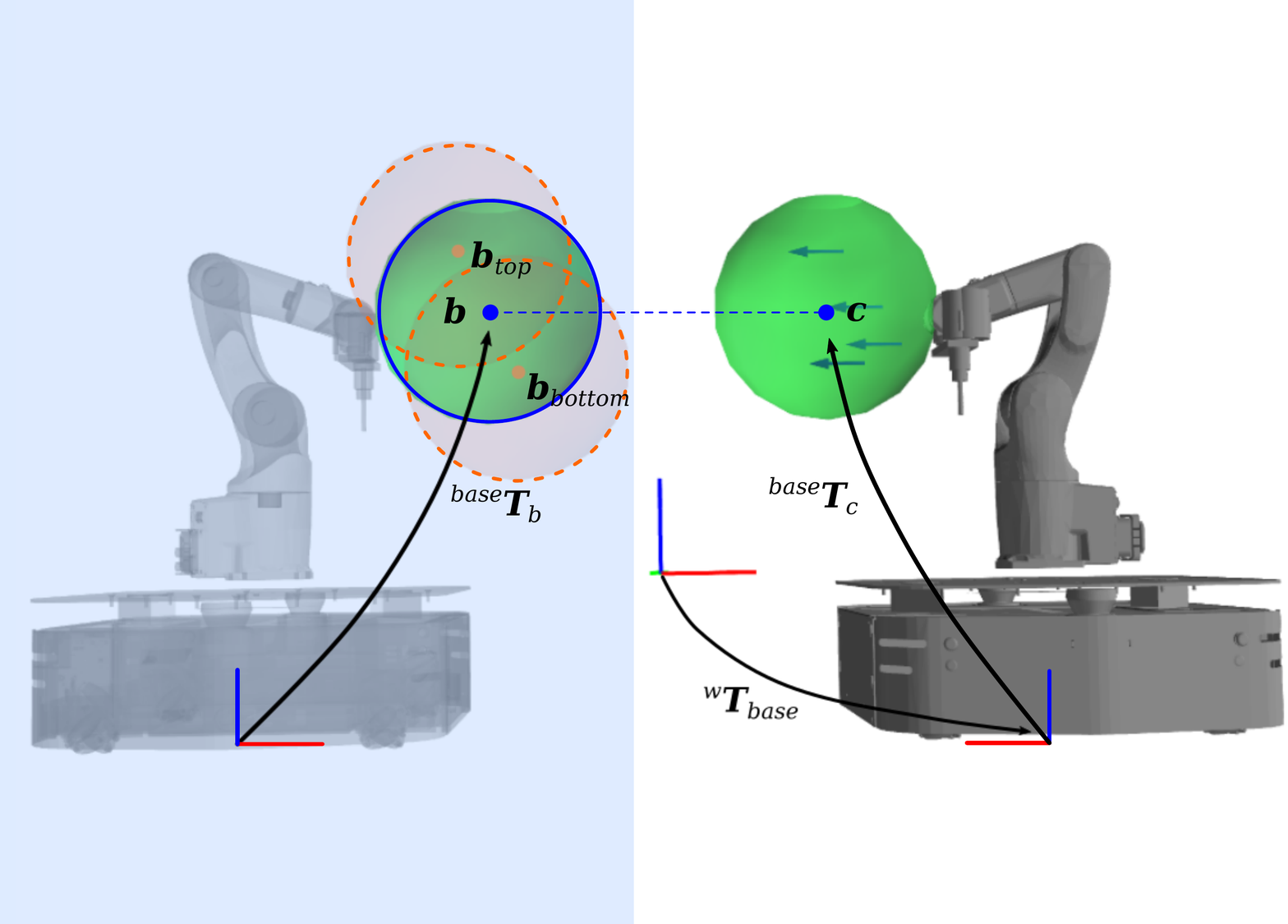}
    \caption{}
    \label{fig:matching}
  \end{subfigure}
  \caption{(a) Illustration of how $\boldsymbol{X}$ is segmented into $c=3$ clusters of $\boldsymbol{X}_1$, $\boldsymbol{X}_2$, and $\boldsymbol{X}_3$ with equal diameter $d$. The red cross represents the center coordinate for $\boldsymbol{X}_1$ cluster. (b) Left: The set of balls $\mathcal{B}$ is defined by the center coordinates $\boldsymbol{b}_\mathrm{top}$ and $\boldsymbol{b}_\mathrm{bottom}$ shown in orange dots. The outer shapes of the top and bottom balls of diameter $d$ are displayed in dashed orange circle. Right: Target cluster $\boldsymbol{X}_i$ is covered by a ball centered at $\boldsymbol{c}_i$ with diameter $d$. Matching: Ball centered at $\boldsymbol{c}_i$ can be matched with the respective ball $B(\boldsymbol{c}_i) \subseteq B$ centered at $\boldsymbol{b}_i$ such that $^{base}\boldsymbol{T}_{\boldsymbol{b}_i} = ^{base}\boldsymbol{T}_{\boldsymbol{c}_i}$.}
  \label{fig:pipeline}
\end{figure}

\begin{enumerate}
\item Find a set of \textit{balls} $\mathcal{B}$, each with diameter $d$, within the robot's reachable workspace. The volume covered by the set of reachable \textit{balls} encloses all possible end-effector positions relative to the robot's base which might be used for visiting targets. In other words, the robot base will always be positioned such that the immediate target will always be strictly located within these set of balls. The details on how $\mathcal{B}$ is obtained is included in Section \ref{part:closed-balls}. Refer to Figure \ref{fig:matching} for illustration of $\mathcal{B}$.

\item Segment targets into workspace clusters. $\boldsymbol{X}$ is segmented into $c$ clusters of $\boldsymbol{X}_i$ for $i \in [1,c]$. $\boldsymbol{X}_i$ is guaranteed to fit geometrically inside a \textit{ball} of center position $\boldsymbol{c}_i$ and diameter $d$. Refer to Figure \ref{fig:covering}.

\item Assign each workspace cluster with one of the reachable \textit{balls} in $\mathcal{B}$. Each cluster \textit{ball} of $\boldsymbol{X}_i$ centered at $\boldsymbol{c}_i$ is assigned to $\mathcal{B}(\boldsymbol{c}_i) \subseteq \mathcal{B}$. Through this matching process, we can immediately obtain the robot's base pose $\boldsymbol{T}^i_{base}$ to visit the target cluster $\boldsymbol{X}_i$. Refer to Figure \ref{fig:matching}.

\item Calculate base tour that minimizes base pose distance in \textit{task-space} between $\boldsymbol{X}_{i}$.
\item Calculate manipulator tour that respects the base tour.
\item Calculate collision-free trajectory based on obtained base and manipulator tour.
\end{enumerate}

\section{Reachable Set of Balls}
\label{part:closed-balls}

We calculate a set of \textit{balls} $\mathcal{B}$, each with diameter \textit{d}, in the robot's reachable workspace. The centers of the \textit{balls} form a continuous straight line in 3D and hence the set can be identified by the two \textit{balls} at the extreme ends.

There is another requirement that $\mathcal{B}$ has to fulfill: any point within $\mathcal{B}$ must be reachable by the end-effector from any directions in $\boldsymbol{R}$. Since it is impossible to check every point in the continuous space covered by $\mathcal{B}$, we apply an approximation condition: Every point in the discretized $\mathcal{B}$ must be reachable from the direction bounding $\boldsymbol{R}$. The reason behind this will be made clear in Section \ref{part:base-pose}.

To achieve that, we construct a \textit{focused kinematic reachability} (fkr) database which involves discretizing the Cartesian workspace into 3D voxels. For each voxel position, we perform Inverse Kinematics (IK) to obtain a set of IK solutions for each direction in $\boldsymbol{R}_{ext} \subseteq \boldsymbol{R}$ where $\boldsymbol{R}_{ext}$ contains the bounding directions for $\boldsymbol{R}$. All voxels containing IK solutions for all directions in $\boldsymbol{R}_{ext}$ are then marked and saved as $\boldsymbol{V}_\mathrm{fkr}$.

\begin{figure}[!t]
  \centering
  \includegraphics[width=\columnwidth]{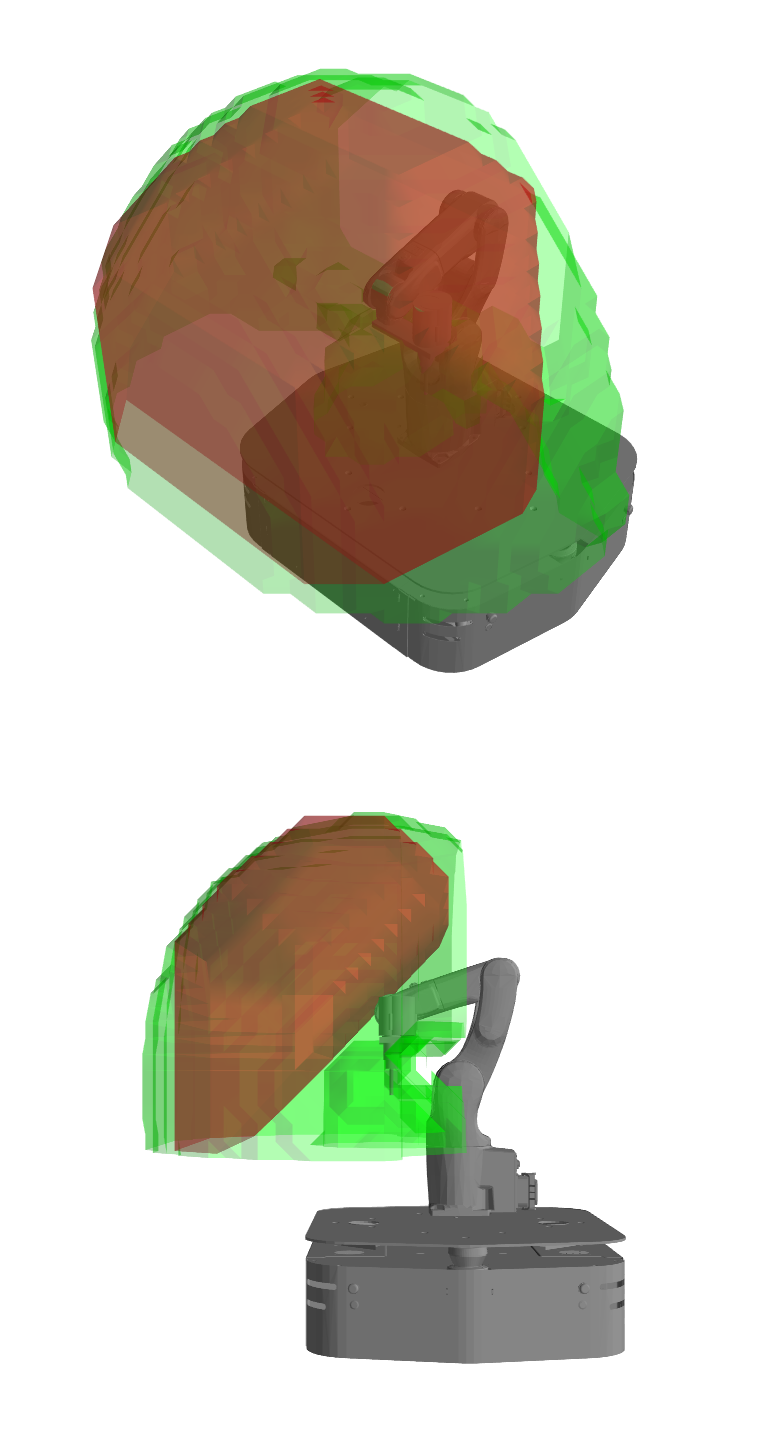}
  \caption{$V_\mathrm{fkr}$, shown in green, marks all the voxels on the top-front quarter of the robotic workspace that are reachable from direction $^{base}\boldsymbol{R}_{ext} = \boldsymbol{R} = [1,0,0]$. The maximal convex subset approximation $M_\mathrm{fkr}$ is shown in red.}
  \label{fig:macs}
\end{figure}

Given that $\boldsymbol{V}_\mathrm{fkr}$ represents the discretized volume that is reachable from the directions bounding $\boldsymbol{R}$, we can now proceed to find the set of balls $\mathcal{B}$ within $\boldsymbol{V}_\mathrm{fkr}$. To make the problem tractable, we can solve this with simple linear programming if we can obtain a maximal convex representation of $\boldsymbol{V}_\mathrm{fkr}$. Finding the largest convex subset is often known as finding the \textit{Maximal Area/Surface Convex Subset} (MACS), \textit{potato-peeling} or \textit{convex skull} problem. More specifically, we are dealing with the \textit{digital}\cite{borgefors2005approximation}\cite{crombez2018peeling} variation given $\boldsymbol{V}_\mathrm{fkr}$ is in integer coordinates. In this paper, we implemented a heuristic for 3D \textit{digital potato-peeling} from \cite{borgefors2005approximation} to obtain a MACS approximation $\boldsymbol{M}_\mathrm{fkr}$. Refer to Figure \ref{fig:macs} for an example of our MACS finding algorithm implementation.

\begin{figure}[t]
  \centering
  \includegraphics[width=\columnwidth]{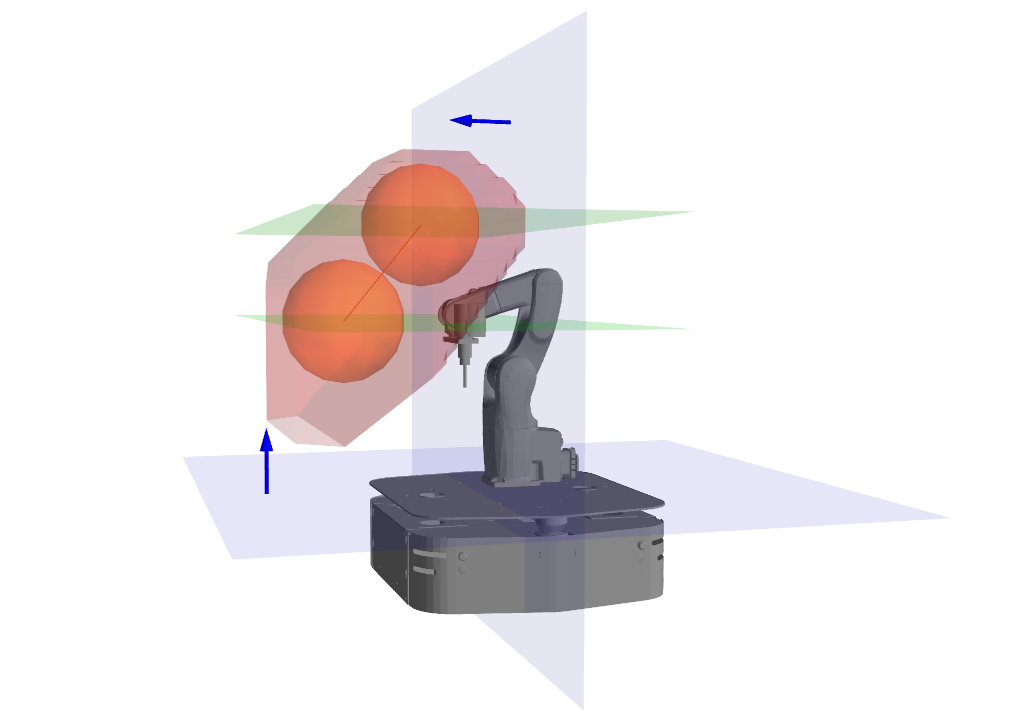}
  \caption{To find the set $\mathcal{B}$, we find the intersection of half-spaces subject to two equality constraints using linear programming. The half-spaces consist of the half-spaces of $M_\mathrm{fkr}$ (red), two collision inequality planes (blue) with vectors showing the normals and two equality planes (green) to cover the minimum and maximum target height respectively. The resulting two balls centered at $\boldsymbol{b}_\mathrm{top}$ and $\boldsymbol{b}_\mathrm{bottom}$ are shown in orange. By convexity, any ball centered along the line between the two balls are inside the $M_\mathrm{fkr}$ half-spaces and hence included in the set $\mathcal{B}$.}
  \label{fig:lp}
\end{figure}

Finally, we find the largest set $\mathcal{B}$ that can be contained inside $\boldsymbol{M}_\mathrm{fkr}$ via linear programming \cite{dantzig1998linear}. Since $\boldsymbol{M}_\mathrm{fkr}$ is convex, we can formulate the problem as simply finding the largest ball diameter $d$ for two balls $\boldsymbol{b}_\mathrm{top}=[b_x^t,b_y^t,b_z^t]$ and $\boldsymbol{b}_\mathrm{bottom}=[b_x^b,b_y^b,b_z^b]$ on extreme ends of $\mathcal{B}$ subject to following constraints as visualized in Figure \ref{fig:lp}:

\begin{align*}
\min \boldsymbol{e}^T\boldsymbol{a}\\
\text{such that}\ \boldsymbol{A}_\mathrm{mfkr}\boldsymbol{a} &\leq \boldsymbol{b}_\mathrm{mfkr}\\
[-1,0,0,0,0,0,1]^T\boldsymbol{a} &\leq x_\mathrm{offset}\\
[0,0,0,-1,0,0,1]^T\boldsymbol{a} &\leq x_\mathrm{offset}\\
[0,0,-1,0,0,0,1]^T\boldsymbol{a} &\leq z_\mathrm{offset}\\
[0,0,0,0,0,-1,1]^T\boldsymbol{a} &\leq z_\mathrm{offset}\\
[0,0,1,0,0,0,0]^T\boldsymbol{a} &= \mathrm{min}(\mathrm{height}(x^i))\\
[0,0,0,0,0,1,0]^T\boldsymbol{a} &= \mathrm{max}(\mathrm{height}(x^i))\\
\end{align*}
\begin{align*}
\text{where}\ \boldsymbol{e}^T &= [0,0,0,0,0,0,-1]\\
\boldsymbol{a} &= [b_x^b,b_y^b,b_z^b,b_x^t,b_y^t,b_z^t,d/2]\\
\boldsymbol{A}_\mathrm{mfkr} &= M_\mathrm{fkr}\ \text{plane coefficients}\\
\boldsymbol{b}_\mathrm{mfkr} &= M_\mathrm{fkr}\ \text{plane offsets}\\
x_\mathrm{offset} &= \text{X-axis collision plane}\\
z_\mathrm{offset} &= \text{Z-axis collision plane}\\
\end{align*}

\section{Workspace Clustering}
\label{part:covering}

In this section we explain how the targets $\boldsymbol{X}$ are segmented into clusters $\boldsymbol{X}_i$ which can be enclosed within a ball of center $\boldsymbol{c}_i$ and diameter $d$. We refer to $\boldsymbol{X}_i$ and the hypothetical enclosing ball interchangeably.

The problem is related to covering of 3D points with balls, which can be formulated as partitioning graph by cliques or \textit{clique covering}. Firstly, we form the undirected graph $\mathcal{G}$ by connecting two nodes (targets) if the Euclidean distance between them is less or equal to $\delta$. Secondly, we run clique clustering algorithm on the undirected graph. Clustering by clique in $\mathcal{G}$ is similar to covering by independent set in the complement graph $\bar{\mathcal{G}}$ which is often referred to as the \textit{graph-coloring} problem. \textit{Graph-coloring} problem is NP-complete but several heuristics exist \cite{kosowski2004classical}\cite{matula1983smallest}\cite{deo2006discrete}. In this paper we implemented greedy coloring of connected sequential breadth-first search variant \cite{kosowski2004classical} as it results in minimum average number of clusters based on our experiment.

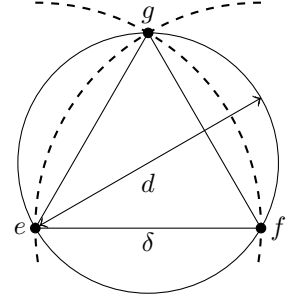
\begin{figure}[t]
\centering
\begin{tikzpicture}[scale=0.6]
\def\x{0};\def\y{0};\def\dlt{5};
\draw[thick,dashed] (\x,\dlt) arc (90:-10:5cm);
\draw[thick,dashed] (\x+\dlt,\dlt) arc (90:190:5cm);
\draw (\x,\y) -- (\x+\dlt,\y);
\def\d{5.773};
\draw[<->,decoration={markings, mark=at position 0 with {\arrow[line width=0.1mm, scale=-1.5]{>}}, mark=at position 1 with {\arrow[line width=0.1mm, scale=1.5]{>}}}, postaction={decorate}] (\x+0.1,\y+0.0588) -- (\x+\dlt,\y+0.5*\d);
\draw (\x,\y) -- (\dlt/2,{\dlt*sqrt(3)/2});
\draw (\x+\dlt,\y) -- (\dlt/2,{\dlt*sqrt(3)/2});
\filldraw[black] (0,0) circle (3pt) node[anchor=east] {$e$};
\filldraw[black] (\x+\dlt,0) circle (3pt) node[anchor=west] {$f$};
\filldraw[black] (\dlt/2,{\dlt*sqrt(3)/2}) circle (3pt) node[anchor=south] {$g$};
\node at (\dlt/2,-0.3) {$\delta$};
\node at (\dlt/2,1) {$d$};
\def\orad{\dlt/sqrt(3)};
\draw (\dlt/2,{\dlt*sqrt(3)/2 - \dlt/sqrt(3)}) circle[radius=(\orad)*1cm];
\end{tikzpicture}
\caption{Diagram containing the minimum circle $O$ with diameter $d$ that covers a maximal clique with maximum edge length $\delta$}
\label{fig:diagram}
\end{figure}

The value of $\delta$ can be calculated such that the resulting clique can be covered by ball of diameter $d$ and is presented here. Let $e$ be the center of a circle with radius $\delta$ and $f$ is a point on the circle's circumference as illustrated in Figure \ref{fig:diagram}. Points $e$ and $f$ are of distance $\delta$ and hence would be connected by an edge if they are nodes in the Graph $\mathcal{G}$. Point $g$ is the intersection point between two circles centered at $e$ and $f$ of similar radius $\delta$ and the three points form a clique. Circle $O$ is smallest circle that circumscribes points $e$,$f$, and $g$. It is clear that any maximal clique that contains the three points will be contained inside $O$ with diameter $d$. The relationship between the ball diameter $d$ and the maximum edge length $\delta$ is then obtained by solving for the circumdiameter of the equilateral triangle $\bigtriangleup{efg}$:

\begin{align*}
\delta = \frac{\sqrt{3}}{2} d 
\end{align*}

\section{Reachable Base Pose}
\label{part:base-pose}

The reachable base pose $\boldsymbol{T}^i_{base}$ to visit each cluster $\boldsymbol{X}_i$ can be analytically determined by matching $\boldsymbol{X}_i$ with one of the reachable balls in $\mathcal{B}$.

$\boldsymbol{X}_i$ centered at $\boldsymbol{c}_i$ is assigned to one of the reachable balls $\mathcal{B}(\boldsymbol{c}_i)$ by height where $\mathcal{B}(\boldsymbol{c}_i) \subseteq \mathcal{B}$. That is, $\boldsymbol{c}_i$ and $\mathcal{B}(\boldsymbol{c}_i)$ share the same z-coordinate.

We can then calculate $\boldsymbol{T}^i_{base}$ by enforcing $\boldsymbol{c}_i$ and $\mathcal{B}(\boldsymbol{c}_i)$ to be physically coincident:

\begin{align}
^{base}\boldsymbol{T}_{\mathcal{B}(\boldsymbol{c}_i)} &= ^{base}\boldsymbol{T}_{\boldsymbol{c}_i}\nonumber
\end{align}

With that we can obtain reachable $\boldsymbol{T}^i_{base}$:

\begin{align*}
^{w}\boldsymbol{T}_{base}^i &= ^{w}\boldsymbol{T}_{\boldsymbol{c}_i} \cdot ^{\boldsymbol{c}_i}\boldsymbol{T}_{base}\\
&= ^{w}\boldsymbol{T}_{\boldsymbol{c}_i} \cdot (^{base}\boldsymbol{T}_{\boldsymbol{c}_i})^{-1}\\
&= ^{w}\boldsymbol{T}_{\boldsymbol{c}_i} \cdot (^{base}\boldsymbol{T}_{\mathcal{B}(\boldsymbol{c}_i)})^{-1}\\
\end{align*}
where $w$ is the world's frame.

\section{Task Sequencing}
\label{part:task-sequencing}
Let $\mathrm{seq}(g(k))$ be a sequence of $k$ configurations in the coordinate space spanned by $g$. The length of the sequence is defined by $\mathrm{length}(\mathrm{seq}(g(k))) = \sum_{n=1}^{k-1} \lVert g(k)_{n+1} - g(k)_n \rVert$. When $g(k)_1 = g(k)_k$, $\mathrm{seq}(g(k))$ forms a tour.

Let $\boldsymbol{t} = [t_x,t_y]$ be the base positions. The algorithm consists in:
\begin{enumerate}
\item Finding a near-optimal base transforms tour $\mathrm{seq}(\boldsymbol{t}(c+2))$ in task-space that minimizes $\mathrm{length}(\mathrm{seq}(\boldsymbol{t}(c+2)))$.
\item Finding the near-optimal target sequence $\mathrm{seq}(\boldsymbol{x}(n+2))$ in task-space given $seq(\boldsymbol{t}(c+2))$.
\item Finding the optimal manipulator $\boldsymbol{q}^* = [q_1,...,q_{n+2}]$ configuration for each target in $\mathrm{seq}(\boldsymbol{x}(n+2))$ such that $\mathrm{length}(\mathrm{seq}(\boldsymbol{x}(c+2)))$ is minimum. Collisions are ignored at this stage.
\item Computing fast collision-free configuration space trajectories on 9-DOF combined base and manipulator configurations based on $\mathrm{seq}(\boldsymbol{t}(c+2))$ and $\mathrm{seq}(\boldsymbol{x}(n+2))$.
\end{enumerate}

Our task sequencing algorithm is inspired by RoboTSP, a fast RTSP solution for fixed-base manipulator \cite{suarez2018robotsp}.

In Step 1, the tour $\mathrm{seq}(\boldsymbol{t}(c+2))$ is calculated which starts at predetermined $\boldsymbol{t}^{\mathrm{start}}$, covers all $\boldsymbol{t}^i$ for all $c$ clusters and ends at $\boldsymbol{t}^{\mathrm{end}}$ where  $\boldsymbol{t}^{\mathrm{start}} =\boldsymbol{t}^\mathrm{end}$. The base station tour is found through finding the approximated solution to the Travelling Salesman Problem (TSP) using the \textit{2-Opt} algorithm \cite{applegate2006traveling}.

In Step 2, the tour $\mathrm{seq}(\boldsymbol{x}(n+2))$ is obtained which starts at predetermined $\boldsymbol{x}_\mathrm{start}$, covers all $n$ number of $\boldsymbol{x}_i$ and ends at $\boldsymbol{x}_\mathrm{end}$. Unlike in previous step, the target tour finding is formulated as a minimum-length Hamiltonian Path problem instead.

\begin{figure}[t]
  \centering
  \includegraphics[width=\columnwidth]{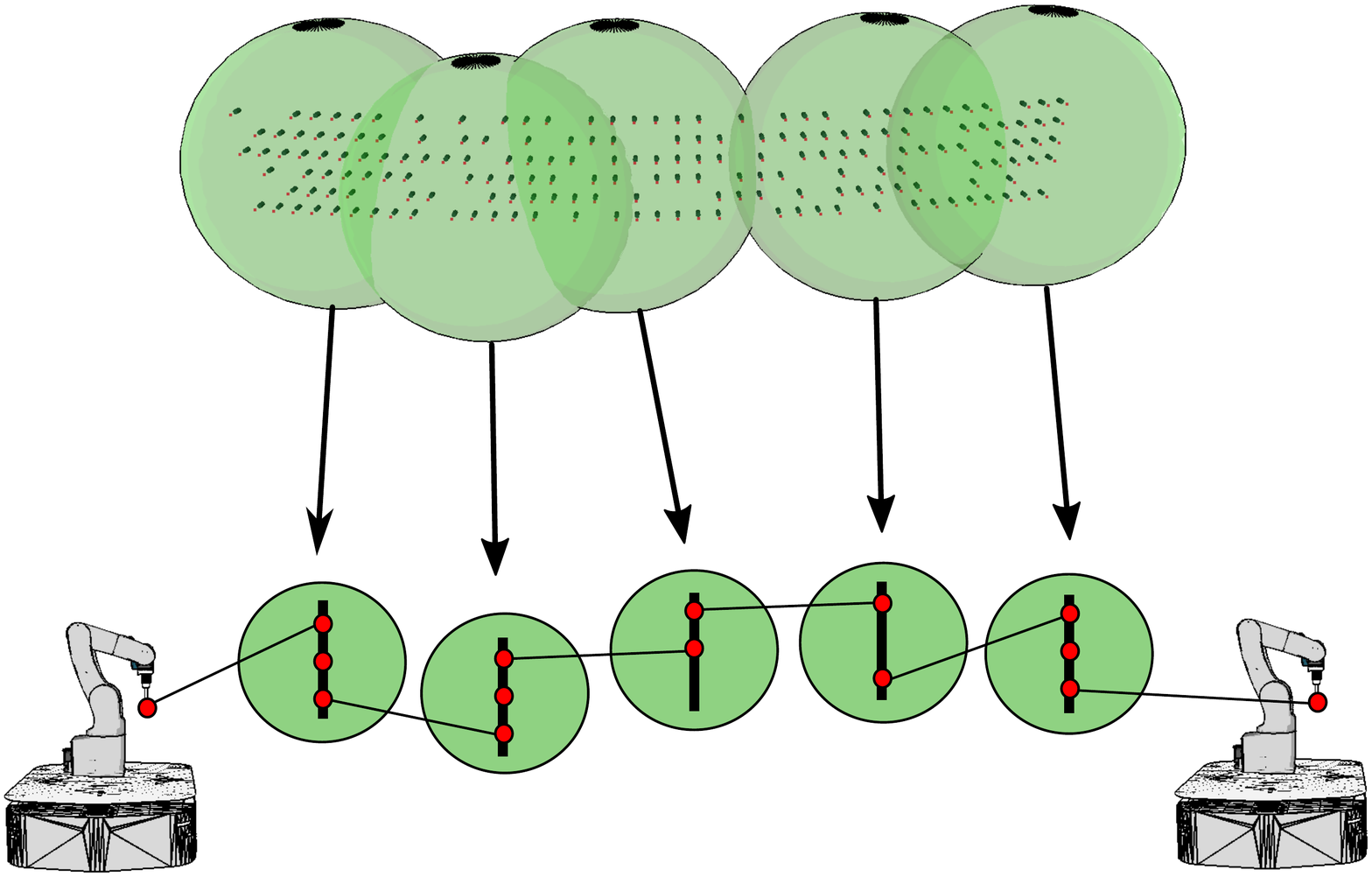}
  \caption{An illustration of \textit{stack-of-clusters}. The clustered targets (top) are ``translated'' and stacked \textit{back-to-back} between a Start and End end-effector position to form $\boldsymbol{x}'_\mathrm{start}$, $\boldsymbol{x}'_1$, $\boldsymbol{x}'_2$, $\boldsymbol{x}'_3$, $\boldsymbol{x}'_4$, $\boldsymbol{x}'_5$ and $\boldsymbol{x}'_\mathrm{end}$ (bottom). Minimum-length Hamiltonian Path is calculated on the ``translated'' clusters to obtain $\mathrm{seq}(\boldsymbol{x}'(7))$, a target tour that respects the base tour sequence $\mathrm{seq}(\boldsymbol{t}(7))$ and minimizes manipulator trajectory distance between base movement.}
  \label{fig:stack-of-clusters}
\end{figure}

This is achieved through virtually translating $\boldsymbol{x}_\mathrm{start}$, $\boldsymbol{x}_1$, ..., $\boldsymbol{x}_n$, $\boldsymbol{x}_\mathrm{end}$ along an axis with distance $h$ between them to form a \textit{stack-of-clusters} as seen in Figure \ref{fig:stack-of-clusters}. We denote $\boldsymbol{x}'$ for the translated $\boldsymbol{x}$. In our implementation, we define $h$ to be:
\begin{align*}
h = &h_\mathrm{scale} \cdot \max_{1 \leq i \leq c} \{{\mathrm{dist}(a,b): a,b \in \boldsymbol{X}_i, a \neq b}\}\\
&\text{where } h_\mathrm{scale} \geq 1
\end{align*}

The minimum-length Hamiltonian Path is calculated on all targets on the \textit{stack-of-clusters} starting from $\boldsymbol{x}'_\mathrm{start}$ and ending at $\boldsymbol{x}'_\mathrm{end}$. The resulting $\mathrm{seq}(\boldsymbol{x}'(n+2))$ is translated back to obtain $\mathrm{seq}(\boldsymbol{x}(n+2))$, which is a target tour sequence that respects the base tour sequence $\mathrm{seq}(\boldsymbol{t}(c+2))$ and minimizes manipulator trajectory distance between base movement.

Similar to \cite{suarez2018robotsp}, in Step 3 we firstly construct an undirected graph of $n+2$ layers following the order in $\mathrm{seq}(\boldsymbol{x}(n+2))$. Each layer $i$, where $i \in [1,n+2]$ contains $v_i$ nodes representing the number of IK solutions of the 6-DOF manipulator for target $i$, resulting in a total of $\sum_{i=1}^{n+2} v_i$ nodes. Note that the first and last nodes are ``Start'' and ``Goal'' nodes respectively such that $v_1 = v_{n+2} = 1$. Next for $i \in [1,n+1]$, we add an edge between each nodes of layer $i$ and layer $i+1$, resulting in a total of $\sum_{i=1}^{n+1} v_i v_{i+1}$ edges. Lastly, we find the shortest path between the ``Start'' and ``Goal'' nodes using a graph search algorithm to find the optimal sequence of IK solutions.

\section{Experiment}
\label{part:experiment}

\begin{figure}[!htp]
  \centering
  \includegraphics[width=\columnwidth,height=0.9\textheight]{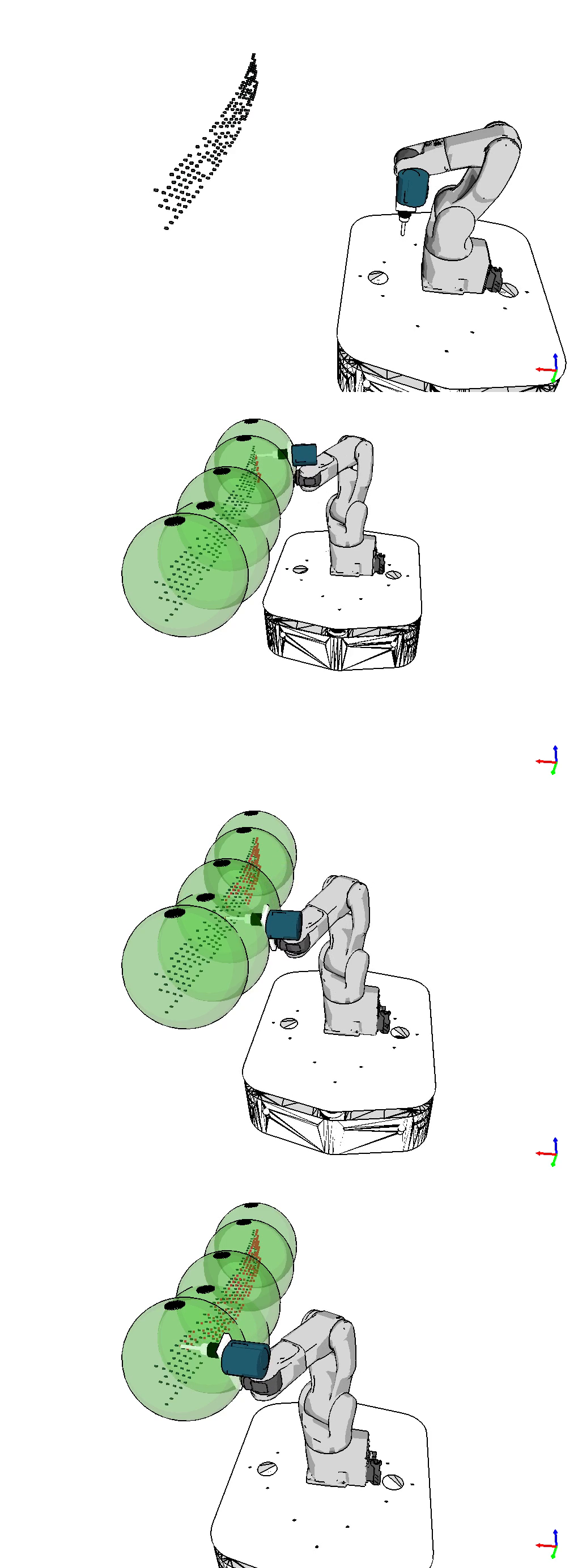}
  \caption{Simulation in OpenRAVE where the mobile manipulator has to visit 183 non-clustered and unsequenced targets. The algorithm found five \textit{workspace clusters} to visit them. The orientations of the targets cover the range within the bounding directions $\boldsymbol{R}_{ext}$ used in building the set of $B$.}
  \label{fig:demo}
\end{figure}

For our experiment we generated the fkr dataset for the top-front quarter grid of the robot's workspace similar to in Figure \ref{fig:macs}. The distance between two workspace voxels is set at 0.04 $\mathrm{m}$. For $\boldsymbol{R_{ext}}$, we consider four direction vectors resembling a pyramid that makes $\theta=10$ degree from the axis that runs along the pyramid's height. The four direction vectors are:
\begin{itemize}
\item $[\mathrm{cos}(\theta), 0, \mathrm{sin}(\theta)]$
\item $[\mathrm{cos}(\theta), 0, -\mathrm{sin}(\theta)]$
\item $[\mathrm{cos}(\theta), \mathrm{sin}(\theta), 0]$
\item $[\mathrm{cos}(\theta), -\mathrm{sin}(\theta), 0]$
\end{itemize}

The drill target directions in simulation are then programmed such that they cover the vector range bounded by values in $\boldsymbol{R_{ext}}$.

Below we provides tables that list the time breakdown of our proposed method. In Table \ref{tab:database}, we show the time required to perform sampling for $V_\mathrm{fkr}$ and subsequently obtaining the maximal digital convex subset $M_\mathrm{fkr}$. We consider these as offline processes as they can be pre-generated and reused on other target distributions with similar robot and target directions requirement. Following that, in Table \ref{tab:cluster} and Table \ref{tab:sequence} we provide the time breakdown for different number of targets for target clustering and target sequencing respectively.

\begin{table}[H]
  \caption{Time breakdown for offline processes}
  \setlength{\tabcolsep}{7pt}
  \centering
  \begin{tabular}{lc}\toprule
    & time (seconds)\\\midrule
    $V_\mathrm{fkr}$ & 225.4\\
    $M_\mathrm{fkr}$ & 2516.1\\
    \bottomrule
  \end{tabular}
  \label{tab:database}
\end{table}

\begin{table}[H]
  \vspace*{5pt}
  \caption{Time breakdown for online target clustering}
  \setlength{\tabcolsep}{9pt}
  \centering
  \begin{tabular}{lc}\toprule
    No of targets & Target clustering in seconds ($\pm$ std)\\\midrule
    183   & 0.104 ($\pm$ 0.01)\\
    268   & 0.209 ($\pm$ 0.04) \\
    2211  & 15.887 ($\pm$ 0.06)\\
    \bottomrule
  \end{tabular}
  \label{tab:cluster}
\end{table}

\begin{table}[H]
  \caption{Time breakdown for online target sequencing}
  \setlength{\tabcolsep}{9pt}
  \centering
  \begin{tabular}{lc}\toprule
    No of targets & Base and hole tour finding in seconds ($\pm$ std)\\\midrule
    183   & 0.919 ($\pm$ 0.01)\\
    268   & 3.804 ($\pm$ 0.06)\\
    \bottomrule
  \end{tabular}
  \label{tab:sequence}
\end{table}

\section{Conclusion}
\label{part:conclusion}

The problem of visiting multiple task points with a mobile manipulator has always been challenging. Compared to its fixed-base counterpart, there are more complexities to deal with when the task space targets span beyond the reachable immediate workspace. Most approaches rely on computationally expensive base poses sampling. To make the problem tractable, they often require the targets to be arbitrarily pre-clustered or pre-sequenced which might not always be possible such as in drilling operations.

In this paper, we introduced a novel approach based on clustering the task point targets first. We presented a a fast and principled way to segment the task point targets into reachable clusters, analytically determine base pose for each cluster and find task sequence to minimize robot's trajectory.

The method here can be easily extended to the continuous case, such as in 3D-printing. However, one needs to enforce the continuity of the manipulator motion, which excludes \textit{IK-switching}. Integrating such a constraint into MoboTSP is the objective of our current research.

\section*{Acknowledgment}
This research was conducted in collaboration with HP Inc. and supported by National Research Foundation (NRF) Singapore and the Singapore Government through the Industry Alignment Fund-Industry Collaboration Projects Grant (I1801E0028).

\bibliography{mtsp}

\begin{thebibliography}{10}

\bibitem{vafadar2018optimal}
S.~Vafadar, A.~Olabi, and M.~S. Panahi, ``Optimal motion planning of mobile
  manipulators with minimum number of platform movements,'' in {\em 2018 IEEE
  International Conference on Industrial Technology (ICIT)}, pp.~262--267,
  IEEE, 2018.

\bibitem{xu2020}
J.~{Xu}, K.~{Harada}, W.~{Wan}, T.~{Ueshiba}, and Y.~{Domae}, ``Planning an
  efficient and robust base sequence for a mobile manipulator performing
  multiple pick-and-place tasks,'' in {\em 2020 IEEE International Conference
  on Robotics and Automation (ICRA)}, pp.~11018--11024, 2020.

\bibitem{shin2003motion}
D.~H. Shin, B.~S. Hamner, S.~Singh, and M.~Hwangbo, ``Motion planning for a
  mobile manipulator with imprecise locomotion,'' in {\em Proceedings 2003
  IEEE/RSJ International Conference on Intelligent Robots and Systems (IROS
  2003)(Cat. No. 03CH37453)}, vol.~1, pp.~847--853, IEEE, 2003.

\bibitem{efe2019}
M.~E. {Tiryaki}, X.~{Zhang}, and Q.~{Pham}, ``Printing-while-moving: a new
  paradigm for large-scale robotic 3d printing,'' in {\em 2019 IEEE/RSJ
  International Conference on Intelligent Robots and Systems (IROS)},
  pp.~2286--2291, 2019.

\bibitem{suarez2018robotsp}
F.~Su{\'a}rez-Ruiz, T.~S. Lembono, and Q.-C. Pham, ``Robotsp--a fast solution
  to the robotic task sequencing problem,'' in {\em 2018 IEEE International
  Conference on Robotics and Automation (ICRA)}, pp.~1611--1616, IEEE, 2018.

\bibitem{zacharias2007capturing}
F.~Zacharias, C.~Borst, and G.~Hirzinger, ``Capturing robot workspace
  structure: representing robot capabilities,'' in {\em 2007 IEEE/RSJ
  International Conference on Intelligent Robots and Systems}, pp.~3229--3236,
  Ieee, 2007.

\bibitem{zacharias2008positioning}
F.~Zacharias, C.~Borst, M.~Beetz, and G.~Hirzinger, ``Positioning mobile
  manipulators to perform constrained linear trajectories,'' in {\em 2008
  IEEE/RSJ International Conference on Intelligent Robots and Systems},
  pp.~2578--2584, IEEE, 2008.

\bibitem{zacharias2009using}
F.~Zacharias, W.~Sepp, C.~Borst, and G.~Hirzinger, ``Using a model of the
  reachable workspace to position mobile manipulators for 3-d trajectories,''
  in {\em 2009 9th IEEE-RAS International Conference on Humanoid Robots},
  pp.~55--61, IEEE, 2009.

\bibitem{vahrenkamp2012manipulability}
N.~Vahrenkamp, T.~Asfour, G.~Metta, G.~Sandini, and R.~Dillmann,
  ``Manipulability analysis,'' in {\em 2012 12th ieee-ras international
  conference on humanoid robots (humanoids 2012)}, pp.~568--573, IEEE, 2012.

\bibitem{vahrenkamp2013robot}
N.~Vahrenkamp, T.~Asfour, and R.~Dillmann, ``Robot placement based on
  reachability inversion,'' in {\em 2013 IEEE International Conference on
  Robotics and Automation}, pp.~1970--1975, IEEE, 2013.

\bibitem{zacharia2005optimal}
P.~T. Zacharia and N.~Aspragathos, ``Optimal robot task scheduling based on
  genetic algorithms,'' {\em Robotics and Computer-Integrated Manufacturing},
  vol.~21, no.~1, pp.~67--79, 2005.

\bibitem{baizid2010genetic}
K.~Baizid, R.~Chellali, A.~Yousnadj, A.~Meddahi, and T.~Bentaleb, ``Genetic
  algorithms based method for time optimization in robotized site,'' in {\em
  2010 IEEE/RSJ International Conference on Intelligent Robots and Systems},
  pp.~1359--1364, IEEE, 2010.

\bibitem{saha2006planning}
M.~Saha, T.~Roughgarden, J.-C. Latombe, and G.~S{\'a}nchez-Ante, ``Planning
  tours of robotic arms among partitioned goals,'' {\em The International
  Journal of Robotics Research}, vol.~25, no.~3, pp.~207--223, 2006.

\bibitem{borgefors2005approximation}
G.~Borgefors and R.~Strand, ``An approximation of the maximal inscribed convex
  set of a digital object,'' in {\em International Conference on Image Analysis
  and Processing}, pp.~438--445, Springer, 2005.

\bibitem{crombez2018peeling}
L.~Crombez, G.~D. da~Fonseca, and Y.~G{\'e}rard, ``Peeling digital potatoes,''
  {\em arXiv preprint arXiv:1812.05410}, 2018.

\bibitem{dantzig1998linear}
G.~B. Dantzig, {\em Linear programming and extensions}, vol.~48.
\newblock Princeton university press, 1998.

\bibitem{kosowski2004classical}
A.~Kosowski and K.~Manuszewski, ``Classical coloring of graphs,'' {\em
  Contemporary Mathematics}, vol.~352, pp.~1--20, 2004.

\bibitem{matula1983smallest}
D.~W. Matula and L.~L. Beck, ``Smallest-last ordering and clustering and graph
  coloring algorithms,'' {\em Journal of the ACM (JACM)}, vol.~30, no.~3,
  pp.~417--427, 1983.

\bibitem{deo2006discrete}
N.~Deo, J.~S. Kowalik, {\em et~al.}, {\em Discrete optimization algorithms:
  with Pascal programs}.
\newblock Courier Corporation, 2006.

\bibitem{applegate2006traveling}
D.~L. Applegate, R.~E. Bixby, V.~Chvatal, and W.~J. Cook, {\em The traveling
  salesman problem: a computational study}.
\newblock Princeton university press, 2006.

\end{thebibliography}
\bibliographystyle{ieeetr}

\end{document}